\documentclass[two column]{article}

\usepackage[english]{babel}
\usepackage{adjustbox}
\usepackage[letterpaper,top=2cm,bottom=2cm,left=2cm,right=2cm]{geometry} 

\usepackage{amsmath}
\usepackage{graphicx}
\usepackage[colorlinks=true, allcolors=blue]{hyperref}
\usepackage{caption}
\usepackage{authblk}
\usepackage{booktabs}
\usepackage{tabularx}
\usepackage{array}
\usepackage{amssymb}
\usepackage{svg}

\title{Large Language Models Illuminate a Progressive Pathway to Artificial Healthcare Assistant: A Review}
\date{}
\author[1,*]{Mingze Yuan}
\author[1,*]{Peng Bao}
\author[2,*]{Jiajia Yuan}
\author[2]{Yunhao Shen}
\author[1]{Zifan Chen}
\author[2]{Yi Xie}
\author[3,7]{Jie Zhao}
\author[2]{Yang Chen}
\author[1,6]{Li Zhang}
\author[2]{Lin Shen}
\author[4-7]{Bin Dong}

\affil[1]{Center for Data Science, Peking University, Beijing, China}
\affil[2]{Department of Gastrointestinal Oncology, Key Laboratory of Carcinogenesis
and Translational Research (Ministry of Education/Beijing), Peking University
Cancer Hospital and Institute, Beijing, China}
\affil[3]{National Engineering Laboratory for Big Data Analysis and Applications, Peking University, Beijing, China}
\affil[4]{Beijing International Center for Mathematical Research, Peking University, Beijing, China}
\affil[5]{Center for Machine Learning Research, Peking University, Beijing, China}
\affil[6]{National Biomedical Imaging Center, Peking University, Beijing, China}
\affil[7]{Peking University Changsha Institute for Computing and Digital Economy, Changsha, China}
\affil[*]{These authors contributed equally}

\begin{document}

\maketitle
\begin{abstract}
With the rapid development of artificial intelligence, large language models (LLMs) have shown promising capabilities in mimicking human-level language comprehension and reasoning. This has sparked significant interest in applying LLMs to enhance various aspects of healthcare, ranging from medical education to clinical decision support. However, medicine involves multifaceted data modalities and nuanced reasoning skills, presenting challenges for integrating LLMs. This paper provides a comprehensive review on the applications and implications of LLMs in medicine. It begins by examining the fundamental applications of general-purpose and specialized LLMs, demonstrating their utilities in knowledge retrieval, research support, clinical workflow automation, and diagnostic assistance. Recognizing the inherent multimodality of medicine, the review then focuses on multimodal LLMs, investigating their ability to process diverse data types like medical imaging and EHRs to augment diagnostic accuracy. To address LLMs' limitations regarding personalization and complex clinical reasoning, the paper explores the emerging development of LLM-powered autonomous agents for healthcare. Furthermore, it summarizes the evaluation methodologies for assessing LLMs' reliability and safety in medical contexts. Overall, this review offers an extensive analysis on the transformative potential of LLMs in modern medicine. It also highlights the pivotal need for continuous optimizations and ethical oversight before these models can be effectively integrated into clinical practice. An accompanying GitHub repository containing latest papers is available at \url{https://github.com/mingze-yuan/Awesome-LLM-Healthcare}.
\end{abstract}
\section{Introduction}
 \begin{figure*}[ht] 
    \centering
    \includegraphics[width=1.0\textwidth]{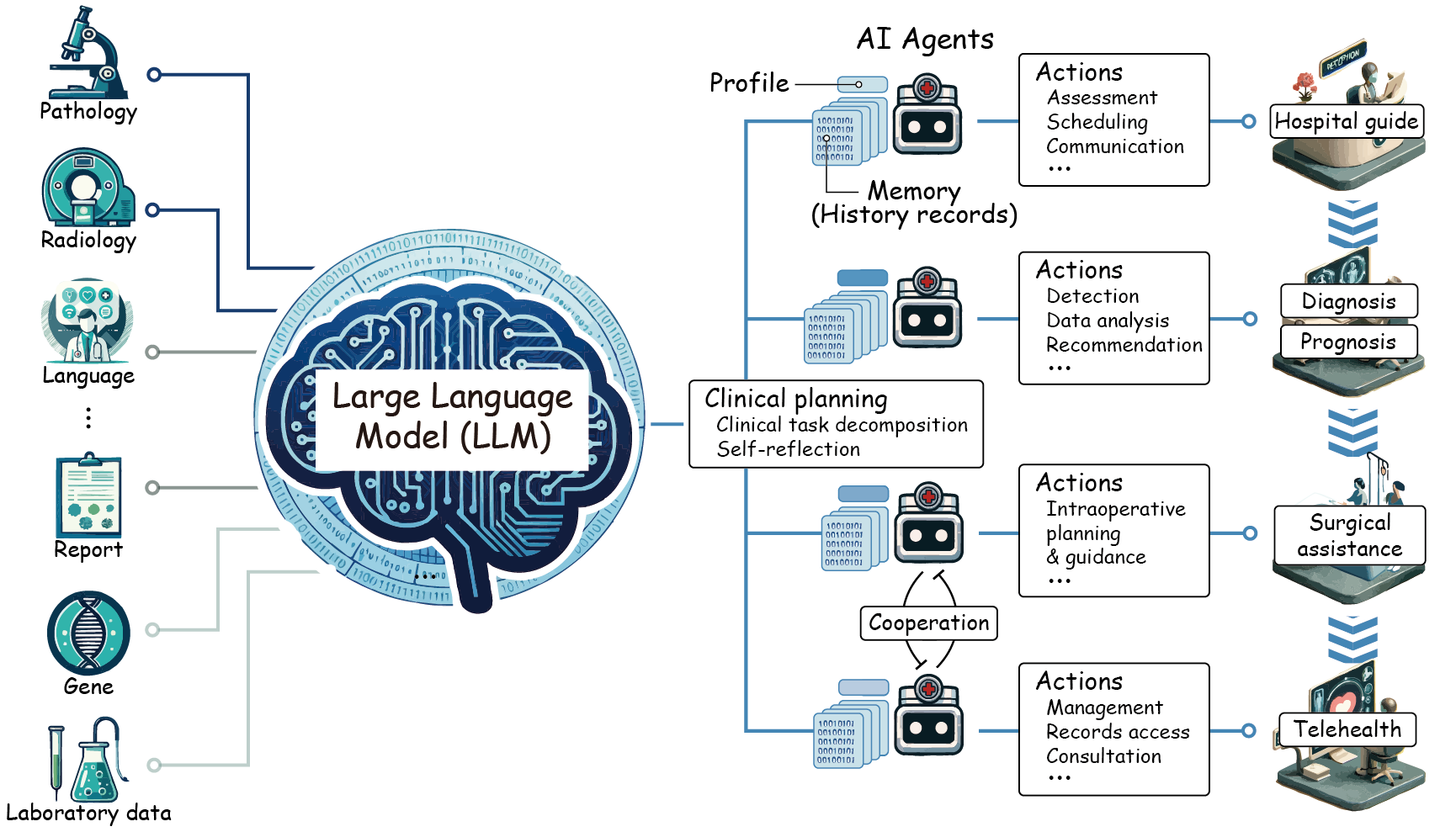}
    \caption{
    Integration of a Large Language Model (LLM) into advanced healthcare support systems. Multimodal medical data, ranging from pathology, radiology to laboratory sources, funnel into the LLM, symbolized by a digital brain. This LLM interacts seamlessly with AI Agents, distinguished by components including profile, planning, memory, and actions. These agents facilitate a range of healthcare procedures, including diagnosis, prognosis, and surgical assistance, underscoring the pivotal role of the LLM in augmenting AI-empowered healthcare systems.
    }
    \label{fig:main}
\end{figure*}

With the onset of the 21st century marked by a staggering growth in artificial intelligence capabilities, we have witnessed groundbreaking advancements and transformations across various industries~\cite{moor2023foundation, ahmed2022artificial, wingstrom2023redefining, lu2022survey,lee2023ai}. Particularly in the medical field, such transformations are even more pronounced~\cite{moor2023foundation,lee2023ai}. However, while AI has unveiled countless new opportunities and possibilities for us, it also sheds light on the profound complexity inherent in medical processes. When considering key stages like diagnosis, treatment, and prognosis, the real-world medical data we grapple with is incredibly diverse and intricate. Doctors, when dealing with this data, not only refer to a vast and complicated body of standard medical knowledge but also need to craft individualized treatment plans based on the unique circumstances of each patient. Furthermore, medical examinations are multimodal, encompassing domains like pathology, radiology, and genomics. Faced with such a scenario, integrating this plethora of data and information to form a coherent and comprehensive diagnosis and treatment strategy is undeniably challenging. Most of the current tools are isolated for single tasks, implying that clinical doctors must engage in more holistic analysis and judgment during decision-making. Hence, there's an urgent need for powerful intelligent assistance tools to aid these physicians. This is precisely what large language models (LLMs), like GPT-4~\cite{OpenAI2023GPT4TR}, offer. Not only can they help doctors consolidate and interpret intricate data, but they also provide insights grounded in extensive knowledge~\cite{haupt2023ai}, thus ensuring more efficient and precise assistance in pivotal stages like diagnosis, treatment, and prognosis~\cite{moor2023foundation,lee2023ai,lee2023benefits}. With the aid of such intelligent tools, we aspire to delve deeper into a patient's genuine situation and make more apt and accurate medical decisions.

In this context, LLMs such as GPT-4~\cite{OpenAI2023GPT4TR}, ChatGPT~\cite{ouyang2022training}, and Claude~\cite{bai2022constitutional} have gradually made their mark in the medical domain. Taking GPT-4 as an example, its exceptional performance in the United States Medical Licensing Examinations (USMLE) has far exceeded the expectations of many experts. Yet, this is only the tip of the iceberg. While the practical application of LLMs in healthcare is still in its early stages, preliminary research has already unveiled their tremendous potential in specialized medical research and potential clinical decision support~\cite{haver2023appropriateness, zhu2023can, bushuven2023chatgpt, xie2023aesthetic, shen2023chatgpt, kothari2023chatgpt, arora2023promise}. Especially in tasks involving the integration of multimodal medical data from pathology, radiology, and genomics, LLMs have exhibited their unique ability for in-depth interpretation and linkage. Of course, their practical effects and values in real medical environments still require further study and validation. With the introduction of these advanced tools, we not only anticipate efficient consolidation of multisource medical data but also expect AI agents~\cite{xi2023rise} to offer support in predictive analysis and patient management for physicians. For instance, AI agents could assist in analyzing patient histories, laboratory results, and radiological data, subsequently providing data-driven diagnostic suggestion~\cite{abbasian2023conversational,zhao2023chatcad+}. Moreover, these tools can further help doctors in choosing the optimal treatment plan from a plethora of options, ensuring patients receive individualized and optimal therapeutic outcomes~\cite{Yuan2023AdvancedPA,haver2023appropriateness}. Through this approach, we can look forward to a medical decision-making process that is not only more scientific but also more systematic, ensuring patients receive the best medical care.

Given the outstanding potential of LLMs in medicine, this paper aims to delve deeply into the recent advances achieved by LLMs in this sector, as illustrated in \autoref{fig:main}. We begin by examining fundamental LLM applications in medicine, spotlighting text-based interactions, and elaborating on both general-purpose and specialized medical LLMs. Subsequently, recognizing the intricate multimodal nature of medicine, we shift our focus to multimodal LLMs, investigating their prowess in integrating diverse data types to enhance diagnostic accuracy and efficacy. Despite significant progress in the field, challenges persist, such as achieving genuine personalization, ensuring continual model updates, and empowering AI with the intricate problem-solving skills required in medicine. Against this backdrop, we detail the applications of LLM-driven autonomous agents in the medical domain, presenting a classification of their varied uses. Ultimately, we summarize LLM evaluation methodologies and engage critically on the prevailing limitations and prospective paths forward.

\begin{table*}[ht]
    \centering
    \small
    \begin{tabular}{m{0.25\textwidth}m{0.7\textwidth}}
        \toprule[1.5pt]
        \textbf{Terminology}  & \textbf{Definition} \\
        \hline
        Large Language Models (LLMs)  & Language models with a large number of parameters, capable of performing a wide array of language tasks. \\\hline
        General-purpose LLMs & LLMs designed to handle a wide variety of tasks without task-specific optimization. \\\hline
        Specialized LLMs & LLMs optimized or fine-tuned for a specific task or domain. \\\hline
        Multimodal LLMs & LLMs capable of understanding and generating content across multiple modalities such as text and images. \\\hline
        AI agents & Advanced systems that act autonomously to carry out tasks or make decisions based on data or environment. \\\hline
        Reinforcement learning with human feedback (RLHF) & A training approach where human feedback is used to optimize the model’s predictions or actions. \\\hline
        Prompt engineering & The practice of crafting and optimizing prompts to effectively instruct a language model. \\\hline
        Zero-shot learning & The ability of a model to generalize to unseen tasks or classes without needing explicit examples during training. \\\hline
        Few-shot learning & The ability of a model to adapt to new tasks or classes with very limited examples. \\\hline
        In-context learning (ICL) & Learning or adapting to new tasks by leveraging context or examples provided during inference. \\\hline
        Fine-tuning & A process of further training a pre-trained model on a specific task to improve its performance. \\\hline
        Instruction tuning & An approach where LLMs undergo additional training using a dataset of instruction-output pairs via supervised learning. \\\hline
        Chain-of-thought prompting& Crafting prompts in a way that guides the model through a multi-step reasoning process. \\\hline
        Reinforcement Learning from AI Feedback (RLAIF) & A reinforcement learning approach where feedback from another AI model is used to guide learning. \\ \hline
        Hallucination & The phenomenon where LLMs may fabricate inconsistent or outright false information.\\
        \bottomrule[1.5pt]
    \end{tabular}
    \caption{This glossary provides concise definitions of crucial terms related to Large Language Models (LLMs) and their applications. It encompasses various types of LLMs, training approaches, learning paradigms, as well as techniques used to optimize and instruct these models.}
    \label{tab:glossary}
\end{table*}

\section{Preliminaries}
Before diving deep into the role of LLMs in the medical domain, it's essential to chart the evolution and highlight key technologies underpinning LLMs. This section serves as a foundation, outlining the development trajectory and salient techniques that have characterized the surge of LLMs in recent years. We offer a glossary pertinent to LLMs and their utilization in \autoref{tab:glossary}.

The emergence of transformer~\cite{vaswani2017attention} has paved the way for the development of large language models (LLMs) in the field of natural language processing (NLP), exemplified by two significant LLMs, namely GPT~\cite{radford2018improving} and BERT~\cite{devlin2018bert}. These LLMs~\cite{radford2018improving,devlin2018bert,radford2019language,brown2020language,ouyang2022training,OpenAI2023GPT4TR} consist of a vast number of learnable parameters, which can easily scale up to billions. They are pre-trained on a large volume of unlabeled corpus using self-supervised learning techniques such as next token prediction~\cite{radford2018improving} and masked language modeling~\cite{devlin2018bert}.

Recent advancements in LLMs, exemplified by models like ChatGPT~\cite{ouyang2022training} and GPT-4~\cite{OpenAI2023GPT4TR}, have showcased outstanding performance as zero-shot or few-shot learners, efficiently summarizing, extracting, and generating text with little to no prompting. The introduction of in-context learning~\cite{brown2020language} has further elevated their capabilities in this domain. Prompt strategies~\cite{radford2019language,lewis2020retrieval,zhou2022least,wei2022chain,yao2023tree}, often used in conjunction with few-shot or zero-shot learning, enhance the performance of LLMs across diverse tasks by conditioning them on a limited set of examples. Standard prompting techniques~\cite{brown2020language} often involve presenting the LLM with a succinct prompt, typically a question or statement, steering the model toward the expected outcome. Techniques like chain-of-thought prompting~\cite{wei2022chain,kojima2022large}, which guides LLMs through a sequence of logical steps towards a solution, and least-to-most prompting~\cite{zhou2022least}, which systematically breaks down intricate problems into more manageable sub-tasks, exemplify the sophisticated strategies employed to harness and optimize the reasoning prowess of these models.

To enable LLMs to understand natural language instructions and perform real-world tasks, researchers have been exploring methods for instruction-tuning of LLMs~\cite{peng2023instruction}. Among these methods, reinforcement learning from human feedback (RLHF)\cite{ouyang2022training} has emerged as a crucial technique for training language models to align with human goals. RLHF has been extensively used to fine-tune state-of-the-art LLMs such as GPT-4\cite{OpenAI2023GPT4TR}, Claude~\cite{bai2022constitutional}, Bard~\cite{google2023bard}, and LLaMA-Chat~\cite{touvron2023llama}. It consists of three interconnected processes: feedback collection, reward modeling, and policy optimization. Specifically, feedback collection involves obtaining evaluations of model outputs from humans and reward modeling aims to train a reward model that mimics these evaluations via supervised learning. Finally, the policy optimization step fine-tunes the language model to produce outputs that garner positive evaluations from the reward model, ensuring alignment with human preferences. A significant challenge with scaling RLHF is the need for copious high-quality human annotations. However, recent research suggests an innovative solution - Reinforcement Learning from AI Feedback (RLAIF)~\cite{lee2023rlaif}. This approach promises performance at par with human-level evaluations, potentially circumventing the scalability challenges inherent to RLHF.

The evolution of LLMs has given rise to the concept of foundation models~\cite{bommasani2021opportunities,moor2023foundation}, which are trained on expansive datasets and demonstrate versatility across diverse downstream applications. Their influence is palpable across various domains, from linguistics~\cite{brown2020language} and vision~\cite{dehghani2023scaling} to other modalities~\cite{borsos2023audiolm}. Intrinsically linked to foundation models is the idea of a generalist model~\cite{moor2023foundation}, wherein a consistent model structure, devoid of fine-tuning, performs commendably across myriad tasks. The aspiration for a unified multitask model~\cite{caruana1997multitask} that proficiently addresses an array of challenges has persisted over time~\cite{collobert2008unified,ruder2017overview}. This ambition is particularly evident in the medical sector~\cite{moor2023foundation}, where contemporary LLMs underscore the feasibility of crafting medical generalist frameworks~\cite{tu2023towards,wu2023towards}.

Alongside the evolution of LLMs, a plethora of in-depth reviews have surfaced. These offer profound understandings of varied facets, encompassing the background, leading-edge technologies, applications, and the inherent challenges in deploying LLMs~\cite{zhao2023survey,yang2023harnessing,chang2023language}. Furthermore, pivotal topics such as harmonizing LLMs with human cognition~\cite{wang2023aligning}, their inherent reasoning capabilities~\cite{huang2022towards}, instruction tuning approaches~\cite{Zhang2023InstructionTF}, augmentation strategies~\cite{mialon2023augmented}, and evaluation methodologies~\cite{chang2023survey} have been encapsulated in recent reviews. Additionally, the burgeoning fields of multimodal LLMs~\cite{Yin2023ASO} and LLM-based agents~\cite{wang2023survey,weng2023prompt,xi2023rise} have been extensively reviewed. 



As we transition into the medical realm, a few studies stand out in their exploration of LLMs and their potential impact. Moor et al.\cite{moor2023foundation} have proposed the concept of a generalist medical AI, though without concrete implementations and empirical evaluations. Rajpurkar and Lungren~\cite{rajpurkar2023current} have provided an insightful review on the evolution, impediments, and prospects of radiological AI models in clinical practice, emphasizing the integral role of LLMs. Qiu et al.~\cite{qiu2023large} have examined the potential impact of expansive AI models, particularly LLMs, in health informatics, pinpointing seven key areas poised for transformation, including molecular biology and drug development. Liu et al.~\cite{liu2023artificial} have highlighted the capacity of artificial general intelligence (AGI) to enhance patient care in radiation oncology through the adept analysis of extensive multimodal clinical data. Moreover, a review by Thirunavukarasu et al.\cite{thirunavukarasu2023large} evaluates LLMs' strengths and limitations, emphasizing their potential to enhance clinical, educational, and research activities. Other recent studies~\cite{li2023chatgpt,sallam2023chatgpt,liu2023utility,clusmann2023future,omiye2023large,he2023survey} have comprehensively examined the potential applications and challenges of LLMs in healthcare. Notably, He et al.~\cite{he2023survey} provided an in-depth overview of current specilized LLMs in the healthcare sector, detailing their training data, methodologies, and performance across three benchmarks.

Despite these invaluable insights, the pathway for forging advanced medical AI frameworks harnessing LLMs remains undefined. In the following sections of this survey, we will delve deeper into the manifold applications and implications of LLMs in the medical domain. Specifically, we will explore the broader applications of general-purpose and specialized LLMs in medicine, focus on the context of multimodal LLMs, and provide an in-depth look into LLM-driven autonomous agents in the medical field. Through this cohesive exploration, our intent is to methodically highlight the transformative potential of LLMs in modern medicine.

\section{LLMs in Medicine} 
Within the medical field, there is a constant pursuit of AI-driven support systems to enhance healthcare delivery. Prior to the advent of LLMs, Watson for Oncology (WFO)~\cite{jie2021meta} emerged as an AI-assisted decision-making tool, collaboratively developed with leading oncologists. This system underwent an extensive training period spanning over four years, drawing upon the National Comprehensive Cancer Network (NCCN) treatment guidelines and amassing over a century’s worth of clinical cancer treatment expertise from the United States. WFO was engineered to propose suitable chemotherapy plans tailored to individual cancer patients. Nevertheless, this initial foray into AI-assisted medicine revealed a profound disconnect between traditional machine learning methodologies and the practical workflow of medical practitioners~\cite{strickland2019ibm}, at times resulting in recommendations that were unsafe and incorrect~\cite{ross2018ibm}.

Large language models, on the other hand, have showcased an enhanced capability for logical reasoning and application of knowledge~\cite{ott2023thoughtsource,singhal2023towards,wei2022chain}. In this section, we focus on the application of basic LLMs in medicine that rely solely on textual information for interaction. Our discussion encompasses two main avenues of research: the first explores the direct application of general-purpose LLMs to medical contexts, whereas the second pursues the development of a specialized medical LLM.

\subsection{Applying General-purpose LLMs to Medicine}
The ascendancy of general-purpose LLMs~\cite{ouyang2022training,OpenAI2023GPT4TR,google2023bard,bai2022constitutional,touvron2023llama} has sparked significant interest in the medical field. Recent literature~\cite{li2023chatgpt,sallam2023chatgpt,liu2023utility} has provided a comprehensive review of ChatGPT’s applications within healthcare and clinical practice. To gauge the capability of these models, researchers frequently resort to benchmark question-answering datasets spanning various medical disciplines, utilizing metrics such as accuracy, recall, and F1 scores for assessment. OpenAI's pivotal study~\cite{OpenAI2023GPT4TR} stands out, showcasing GPT-4's commendable performance on academic and professional tests tailored for an erudite audience. The results pointed to GPT-4's distinguished aptitude in subjects like the Uniform Bar Exam and GRE. Furthermore, Microsoft's independent analysis~\cite{nori2023capabilities} placed GPT-4 above the USMLE, an exhaustive medical residents' examination, marking a notable improvement from its predecessor, ChatGPT, which merely matched college-level performance on the USMLE~\cite{gilson2023does,kung2023performance}. This progression epitomizes the brisk evolution of LLMs in medical settings.

Subsequent research accentuates the adaptability of general-purpose LLMs across diverse medical subspecialties, ranging from oncology~\cite{haver2023appropriateness,zhu2023can,sorin2023large} to emergency medicine~\cite{bushuven2023chatgpt}, medical aesthetics~\cite{xie2023aesthetic}, radiology~\cite{shen2023chatgpt}, ophthalmology~\cite{mihalache2023performance,hu2023can}, surgery~\cite{humar2023chatgpt}, and nursing~\cite{kothari2023chatgpt}. These inquiries typically gauge an LLM's domain-specific expertise using carefully curated questions. For example, Hu et al.~\cite{hu2023can} evaluated GPT-4 by progressively introducing information on select ophthalmic conditions, simulating patient and physician interactions. Experienced ophthalmologists subsequently assessed the model's outputs, underscoring GPT-4's potential utility in both patient referrals and medical training. Additionally, Brin et al.~\cite{brin2023comparing} scrutinized both ChatGPT and GPT-4's capabilities in handling USMLE questions centered on communication nuances, ethics, and empathy, finding a noteworthy capacity for empathy and professionalism in AI.

In the realm of knowledge retrieval and dissemination, LLMs emerge as potent instruments, serving not only as invaluable repositories of medical information but also as sophisticated educators. These models afford healthcare professionals immediate access to contemporaneous medical data by meticulously analyzing a plethora of scientific journals, research articles, and clinical protocols. This analysis furnishes pertinent and timely details~\cite{jin2023retrieve,sallam2023chatgpt} pertaining to disease processes~\cite{biswas2023role,rahsepar2023ai}, therapeutic approaches~\cite{cheng2023chatgpt,carlbring2023new,Yuan2023AdvancedPA}, and drug interactions~\cite{he2023chat,blanco2023role}. Such insights can be especially valuable in assisting with the diagnosis of rare diseases~\cite{sun2023gpt}, which often presents challenges for clinicians. Furthermore, LLMs have the dexterity to democratize medical knowledge through online medical consultation~\cite{haupt2023ai,sun2023gpt,zhu2023can,howard2023chatgpt,xie2023aesthetic,yeo2023assessing}, ensuring widespread availability while simultaneously offering customization to cater to individual prerequisites, potentially impacting telemedicine~\cite{liu2023utility,shea2023use}.

The integration of LLMs in medical research and writing, as highlighted in various studies~\cite{biswas2023chatgpt,sallam2023chatgpt,arora2023promise,li2023chatgpt,ghim2023transforming,peng2023study}, significantly enhances the efficiency, equity, and applicability of research endeavors. These models streamline experimental design, ensure the preservation of patient confidentiality through effective anonymization of medical records, and augment the available medical text data for training purposes. Notably, they facilitate the swift collection, processing, and sophisticated analysis of disease-specific data, fostering more comprehensive and insightful research initiatives. Clinical trials, an essential component of medical research, benefit immensely from LLMs, as they address challenges related to patient-trial matching and trial planning~\cite{woo2019ai,hamer2023improving,jin2023matching,white2023clinidigest,wang2023autotrial,ghim2023transforming}. A detailed review by Ghim et al.~\cite{ghim2023transforming} delves into the transformative potential of LLMs within clinical trials, identifying five key areas for imminent implementation: improved patient-trial matching, streamlined clinical planning, advanced free text narrative analysis for coding and classification, assistance in technical trial planning, and the facilitation of informed consent through LLM-powered chatbots. In particular, Jin et al.~\cite{jin2023matching} demonstrated the capability of LLMs, through their proposed TrialGPT system, to aid patients and referral physicians in selecting appropriate clinical trials from a vast array, validating the explanatory prowess and invaluable contribution of LLMs to medical research on three public cohorts encompassing 184 patients and 18,238 annotated clinical trials.

In the context of clinical workflow, LLMs can significantly mitigate the substantial burden shouldered by healthcare professionals by autonomizing the documentation of patient information, clinical observations, and test reports~\cite{arora2023promise}. This automation does more than merely streamline the process; it enhances both the accuracy and the thoroughness of the clinical documentation. For example, LLMs have been efficaciously utilized to summarize radiology reports~\cite{fink2023potential}, providing a prototype for analogous applications in various domains~\cite{haver2023appropriateness}, including using ChatGPT to write patient clinic letters~\cite{ali2023using}. Besides, the deployment of LLMs in clinical decision support is markedly beneficial, offering insightful recommendations pertaining to medication regimens, suggesting suitable imaging services grounded in clinical presentations, and enabling the astute diagnosis of diseases from comprehensive clinical data sets~\cite{shen2023chatgpt}. When synergistically integrated with other diagnostic instruments, such as medical imaging tools, LLMs proffer a more holistic perspective of patient health. Moreover, by analytically examining data from analogous cases, LLMs can prognosticate patient outcomes, thus assisting both healthcare professionals and patients in navigating toward enlightened treatment decisions.

Beyond such application-oriented evaluations, comparative studies have explored methodological optimizations in LLMs for medical applications, underscoring the critical role of task-adapted prompting for optimal performance. Wang et al.~\cite{wang2023large} assessed the prowess of state-of-the-art LLMs, including GPT-3.5~\cite{ouyang2022training}, GPT-4~\cite{OpenAI2023GPT4TR}, and Bard~\cite{google2023bard}, across an array of clinical linguistic tasks, leveraging diverse learning and prompting techniques~\cite{wei2022chain,kojima2022large,brown2020language}. Parallel to this, Yuan et al.~\cite{Yuan2023AdvancedPA} spotlighted GPT-4's proficiency in intricate clinical assignments, particularly in tumor treatment planning, by employing sophisticated prompting methods. Liu et al.'s study~\cite{liu2023evaluating} shed light on LLMs' efficiency in radiology, underlining the variance in results depending on specific shot configurations. Moreover, Tang et al.~\cite{tang2023evaluating} critiqued zero-shot LLMs' abilities in condensing medical evidence, revealing occasional discrepancies in their summaries, thus highlighting the ongoing need for rigorous monitoring and perpetual enhancements.

The potential of general-purpose LLMs is undoubtedly transformative. Yet, when it comes to their integration into the medical sector, numerous challenges arise. Specific tasks and limited question domains can introduce selection bias~\cite{hu2023can}. Furthermore, the model's efficacy often hinges on the design of the prompt, which might not always be intuitive for users~\cite{Yuan2023AdvancedPA}. Notable commercial LLMs, such as GPT-4 and Claude-2, encounter difficulties when being integrated into clinical workflows. These difficulties range from ensuring HIPAA compliance and safeguarding patient privacy to obtaining necessary IRB approvals, particularly when there's a possibility of patient data being transferred to external servers. From an ethical standpoint, it's worth noting that commercial LLMs like ChatGPT might be restricted from delivering medical diagnoses or drug recommendations~\cite{zhang2023huatuogpt}. Unlike physicians who delve deeper into patients' complaints, models like ChatGPT might provide more generic answers~\cite{zhang2023huatuogpt}. Moreover, they struggle to incorporate the latest health insights, leading to potentially outdated responses. A crucial observation is that some models exhibit deficiencies in specialized medical knowledge, a point emphasized by Antaki et al.~\cite{antaki2023evaluating}. One reason for this might be that these LLMs primarily learn from clinical guidelines and research papers—sources that usually reflect controlled environments rather than the nuanced realities of everyday clinical practice. Such challenges amplify the pressing demand for LLMs that are specialized for medical contexts and which are adept at navigating the intricate regulatory landscape~\cite{mao2023gpteval}.

\subsection{Developing Specialized Medical LLMs}
Specialized medical LLMs~\cite{zhang2023huatuogpt,yang2022large,singhal2023large,singhal2023towards,li2023chatdoctor,wang2023huatuo,xiong2023doctorglm,wu2023pmc,chen2023bianque,wang2023clinicalgpt,liu2023radiology,jiang2023health} are meticulously crafted to address the intricate needs of the healthcare sector, with endeavors centered on either developing entirely new LLMs pre-trained with a healthcare-centric focus or refining existing models to bolster their efficacy within medical contexts. The advent of this category is rooted in research findings that highlight the limitations of general-domain LLMs, particularly when tasked with healthcare-specific challenges, as they tend to grapple with domain shifts~\cite{jin2023matching,nori2023capabilities,hu2023can,antaki2023evaluating}. These findings also suggest that merely depending on prompt engineering might not yield substantial enhancements in their performance regarding healthcare-specific applications~\cite{Yuan2023AdvancedPA}.

One prevalent approach to constructing these specialized medical LLMs involves fine-tuning a base LLM on medical dialogue or datasets tailored to specific instructions. A primary concern with generic LLMs in the realm of medicine is the potential discord between their initial training goals and end-user expectations~\cite{zhang2023huatuogpt}. While LLMs traditionally aim to minimize prediction errors across diverse datasets, users desire models that deliver accurate and safe results. Instruction tuning~\cite{peng2023instruction} emerges as a solution, refining LLMs with paired sets of user directives and expected outcomes to better align with users' anticipations.

A noteworthy breakthrough in the field is Google's Med-PaLM~\cite{singhal2023large}. This system stands out as the first AI to excel in USMLE-style inquiries, consistently delivering insightful responses to general health queries. To ensure comprehensive evaluations, the researchers introduce MultiMedQA, an all-encompassing benchmark that amalgamates six previous medical QA datasets across different specialties. Furthermore, they introduce a novel dataset, HealthSearchQA. Building on this foundation, Flan-PaLM is derived through instruction tuning on PaLM~\cite{Chowdhery2022PaLMSL} and subsequently evaluates using MultiMedQA. Impressively, Flan-PaLM outperforms the prior leading model by a margin of 17\% in accuracy. This achievement underscores the value of leveraging real-world medical data. Additionally, human evaluators identifies key areas for enhancement. In response to this feedback, researchers develops the ``instruction prompt tuning" methodology. This innovation paves the way for the launch of Med-PaLM 2 in March 2023~\cite{singhal2023towards}, which boasts an admirable accuracy rate of 86.5\% for USMLE-style questions.

However, high-performing models like GPT-4~\cite{OpenAI2023GPT4TR} and Med-PaLM 2~\cite{singhal2023towards} remain proprietary, hampering their private use in this sensitive field. To circumvent this, efforts have been directed at refining open-source LLMs~\cite{liu2023evaluating}, exemplified by Meta's LLaMA~\cite{touvron2023llama}. ChatDoctor~\cite{li2023chatdoctor} stands out in this respect, having fine-tuned LLaMA using extensive patient-doctor dialogues from a popular online medical platform. It further enhances the model with a self-guided information retrieval feature, enabling access to real-time online resources and trusted offline medical databases. This approach markedly amplifies the model's capability to discern patient requirements and offer reliable counsel. Similarly, Radiology-GPT~\cite{liu2023radiology} utilizes the Alpaca instruction-tuning framework~\cite{alpaca} to create a radiology-centric LLM. The model is tailored to generate diagnostic ``Impression" narratives based on provided ``Findings" data. While Radiology-GPT performs comparably to ChatGPT in understandability and even surpasses it in coherence, it lags slightly in relevance. Importantly, this underscores the potential for crafting domain-specific LLMs that cater to distinct medical niches, while upholding rigorous privacy standards.

When applied to unstructured textual data, such as Electronic Health Records (EHR), a unique strength of specialized LLMs emerges. Specialized LLMs often outperform traditional structured predictive models due to their inherent flexibility. Jiang et al.~\cite{jiang2023health} demonstrate that unstructured clinical notes from EHRs can facilitate the training of clinical language models. These models can then serve as versatile clinical predictive engines, allowing for streamlined development and deployment. Specifically, their approach, a specialized LLM named NYUTron, is pre-trained on a decade's worth of inpatient clinical notes. It is then fine-tuned for a broad spectrum of clinical and operational predictive tasks, delivering significant enhancements over conventional methods.

\section{Multimodal LLMs in Medicine}
\renewcommand{\arraystretch}{1.5}
\begin{table*}[t]
\centering
\small
\begin{tabularx}{\textwidth}
{>{\raggedright\arraybackslash}m{2.5cm} >{\raggedright\arraybackslash}m{2.5cm} >{\raggedright\arraybackslash}m{2.7cm} >{\raggedright\arraybackslash}m{2.5cm} >{\centering\arraybackslash}m{2cm}
>{\raggedright\arraybackslash}m{4cm}}
\toprule[1.5pt]
\textbf{Model} & \textbf{Modality} & \textbf{Task} & \textbf{Base Model} & \textbf{Sample Size} & \textbf{Data Source}\\ \hline
PathAsst~\cite{Sun2023PathAsstRP} & Pathology & Pathological diagnosis & PLIP~\cite{Huang2023AVF}, Vicuna-13B~\cite{zheng2023judging}& 142K&  Open-source (PubMed, books, websites), in-house\\ \hline
BiomedGPT~\cite{zhang2023biomedgpt} & Radiology, pathology & VQA, image captioning & OFA~\cite{wang2022ofa}&$\sim$184M & Open-source datasets \\ \hline
PMC-VQA~\cite{Zhang2023PMCVQAVI} & Radiology, pathology, microscopy, \textit{etc.} & VQA & PMC-CLIP~\cite{lin2023pmc}, PMC-LLaMA~\cite{wu2023pmc} & 227K & Open-source (PubMed), self-instruction\\ \hline
LLaVA-Med~\cite{li2023llava} & Radiology, pathology & VQA & LLaVA~\cite{liu2023visual}, CLIP~\cite{radford2021learning}& 600K & Open-source (PubMed~\cite{zhang2023large}), self-instruction \\ \hline
XrayGPT~\cite{Thawakar2023XrayGPTCR} & X-ray & VQA & MedCLIP~\cite{wang2022medclip}, Vicuna~\cite{zheng2023judging} & 217K & Open-source datasets (MIMIC-CXR~\cite{johnson2019mimic}, OpenI~\cite{demner2016preparing}) \\ \hline
CephGPT-4~\cite{Ma2023CephGPT4AI} & Dental imaging & Orthodontic measurement \& diagnostic & MiniGPT-4~\cite{zhu2023minigpt}, VisualGLM~\cite{du2022glm} & $\sim$60K & Patient dialogues, real clinical case samples \\ \hline
Med-PaLM M~\cite{tu2023towards} & Radiology, pathology, mammography, genomics, dermatology, \textit{etc.} & QA, VQA, report summarization \& generation, genomic variant calling, \textit{etc.} & PaLM-E~\cite{Driess2023PaLMEAE} & $>$1M& Open-source (12 de-identified datasets) \\ \hline
Med-Flamingo~\cite{Moor2023MedFlamingoAM} & Radiology, pathology, \textit{etc.} & VQA & OpenFlamingo~\cite{Awadalla2023OpenFlamingoAO} &1.3M& Open-source (textbooks, PubMed~\cite{lin2023pmc}) \\ \hline
RadFM~\cite{wu2023towards} & Radiology & VQA, disease diagnosis, report generation & ViT~\cite{Dosovitskiy2020AnII}, PMC-LLaMA~\cite{wu2023pmc}&16M & Open-source (a collection of existing datasets), self-instruction \\ \hline
BioMedGPT~\cite{luo2023biomedgpt} & Molecule, protein & QA (medical, molecule, protein) & LLaMA2-7B-Chat~\cite{touvron2023llama} & $\sim$2.3M & Open-source (literature~\cite{lo2019s2orc}, PubChem) \\ \hline
HeLM~\cite{Belyaeva2023MultimodalLF} & Individual-specific (e.g., lab values) & Disease risk estimation & PaLM-E~\cite{Driess2023PaLMEAE} & $\sim$17K & Open-source (UK Biobank) \\ 
\bottomrule[1.5pt]

\end{tabularx}
\caption{A summary of multimodal LLMs in medicine. This table does not include the multitude of open-source projects available on Github~\cite{liu2023evaluating}. The ``Sample Size'' column denotes the number of training samples, such as image-text pairs. ``VQA'' stands for Visual Question Answering, and ``QA'' stands for Question Answering.}
\label{tab:multimodal}
\end{table*}

Medicine inherently involves multiple data modalities, including text, images, genomics, and more. A recent review on multimodal biomedical AI~\cite{acosta2022multimodal} has explored the opportunities for multimodal datasets in healthcare, and then discussed the key challenges and promising strategies for overcoming them. However, unimodal LLMs still lack the ability to perceive visual modalities such as MRI and handle complex unstructured data (\textit{e.g.}, gene screening status), which limits their utilization for real-world medical scenarios. 

In this section, we focus on multimodal LLMs (MLLMs) in medicine~\cite{wang2023XrayGLM,Sun2023PathAsstRP,zhang2023biomedgpt,Zhang2023PMCVQAVI,li2023llava,Thawakar2023XrayGPTCR,Ma2023CephGPT4AI,tu2023towards,Moor2023MedFlamingoAM,wu2023towards,wu2023can,zhou2023path}, which integrate various modalities into LLMs for diagnostic functions, as listed in \autoref{tab:multimodal}. Here, we specifically examine the taxonomy of utilized data modality, primarily including imaging and intricate unstructured data~\cite{Theodoris2023TransferLE,huang2023chatgpt,Belyaeva2023MultimodalLF,luo2023biomedgpt}, and proceed with their methodologies for data collection and modality fusion.

\subsection{A Taxonomy of Data Modality Usage}
In the field of medicine, multimodal LLMs can be classified based on the data modality they utilize. They primarily fall into two main categories: imaging, which is the most prominent, and other complex unstructured data types, such as genomic sequences, time-series data, and audio recordings. 
\subsubsection{Imaging}
Existing research on multimodal LLMs in medicine, as evidenced by numerous studies~\cite{zhang2023biomedgpt,Sun2023PathAsstRP,Zhang2023PMCVQAVI,li2023llava,Ma2023CephGPT4AI,tu2023towards,Moor2023MedFlamingoAM,wu2023towards,xu2023elixr}, predominantly focuses on exploiting imaging data. The overarching goal is to devise a universal and adaptive model compatible with various imaging modalities and assignments. BiomedGPT~\cite{zhang2023biomedgpt} exemplifies this by offering a versatile AI model for medical applications, which integrates diverse modalities, from CT images to clinical notes. Uniquely, BiomedGPT encapsulates information from various input sources into a shared multimodal lexicon suitable for many tasks. It uniformly employs a sequence-to-sequence paradigm throughout both pretraining and finetuning phases. Furthermore, task directives are seamlessly integrated into inputs as plain text, eliminating the need for supplemental parameters. After rigorous testing on multiple biomedical datasets and tasks, BiomedGPT not only demonstrates its ability to effectively disseminate knowledge across tasks but also matches or outperforms dedicated models optimized for specific datasets or modalities. Its strength is most apparent in vision-language assignments like image captioning and visual question answering, where it sets new benchmarks in performance.

The Med-PaLM Multimodal (Med-PaLM M)~\cite{tu2023towards} further exemplifies a cohesive model tailored to interpret a spectrum of biomedical data types, managing diverse tasks using a consistent set of model weights. Addressing the lack of comprehensive multimodal medical benchmarks, it introduced MultiMedBench, an inclusive open-source multimodal medical benchmark. This benchmark covers language, medical imaging, and genomics, encompassing a wide range of tasks. These include question answering, visual question answering, medical image categorization, radiology report creation and summarization, and genomic variant identification. With this foundation, Med-PaLM M introduces a versatile multimodal sequence-to-sequence architecture capable of smoothly integrating diverse biomedical data. The model's universal language decoder provides inherent flexibility, enabling it to handle a variety of biomedical tasks within a unified generative framework. Impressively, even without task-specific fine-tuning, Med-PaLM M matches or surpasses dedicated models across several MultiMedBench tasks. Beyond just performance metrics, the model demonstrates intuitive medical reasoning, adaptability to novel concepts and responsibilities, and effective knowledge transfer. This underscores its vast potential, especially in areas with limited biomedical data."

RadFM~\cite{wu2023towards} serves as a foundational model for radiology. It curates an extensive multimodal dataset, MedMD, boasting roughly 16M medical scans. This dataset includes 15.5M 2D scans and 180K 3D radiological images, each accompanied by textual narratives, such as radiology reports, visual-language instructions, or vital diagnostic labels. Distinctively, RadFM operates as a text-generation model conditions on visual inputs, adeptly merging natural language with 2D or 3D medical imagery. Its output is primarily in the form of natural language, catering to a diverse set of medical tasks. Additionally, RadFM presents a comprehensive radiology benchmark, capturing a spectrum of clinical duties like disease identification, report drafting, and visual question answering across various radiological modalities and anatomical sectors. Also within the field of radiology, ELIXR~\cite{xu2023elixr} employs a language-aligned image encoder and skillfully integrates it with a stable LLM, specifically PaLM 2~\cite{anil2023palm}, enabling it to handle a diverse set of tasks. This lightweight adapter architecture is trained on images paired with their corresponding free-text radiology reports, sourced from the MIMIC-CXR dataset~\cite{johnson2019mimic}. This configuration underscores the potential of LLM-aligned multimodal models, demonstrating how the combination of chest X-rays with relevant radiology reports can address numerous medical tasks, including visual question answering and radiology report quality assessment.

The recent advancement in the GPT-4 series, GPT-4V~\cite{yang2023dawn}, has introduced support for multimodal inputs, garnering immediate attention due to its potential effectiveness. Wu et al.~\cite{wu2023can} conduct an in-depth evaluation of GPT-4V's performance in multimodal medical diagnostics, encompassing 17 human body systems and employing images from 8 different modalities common in daily clinical practice. The researchers scrutinize GPT-4V’s capacity to handle a variety of clinical tasks, assessing its proficiency both with and without patient history, and spanning activities such as imaging modality and anatomy recognition, disease diagnosis, report generation, and disease localization. While the model excels at distinguishing between medical modalities and identifying anatomical structures, it faces challenges in disease diagnosis and producing detailed medical reports. This study underscores that, despite considerable progress in computer vision and natural language processing within large multimodal models, there remains a considerable gap before these tools can be effectively integrated into real-world medical applications and clinical decision-making. However, it is crucial to acknowledge the limitations of this study, as real clinical settings primarily use 3D DICOM formatted radiological images, whereas GPT-4V can process only up to four 2D images simultaneously, necessitating the selection of 2D key slices or small patches for pathology.

\subsubsection{Other Unstructured Data}
In medical care, clinicians frequently analyze a variety of data types, not limited to medical imaging, but also including clinical notes, lab tests, vital signs, genomics, and other observational metrics. Therefore, effectively deciphering this vast, unstructured data is essential for the integration of multimodal LLMs in healthcare~\cite{luo2023biomedgpt,huang2023chatgpt,Belyaeva2023MultimodalLF}. A recent perspective by Moor et al.~\cite{moor2023foundation} underscores the potential of foundational LLMs, which not only incorporate extensive medical knowledge but also adeptly handle intricate unstructured data. Inspired by the transfer learning paradigm of LLMs, Theodoris et al.~\cite{Theodoris2023TransferLE} propose Geneformer. This model is pre-trained on a substantial corpus of approximately 30 million single-cell transcriptomes, allowing for context-specific predictions in network biology scenarios, especially when data is limited. 

In a recent review by Huang et al.~\cite{huang2023chatgpt}, the potential applications of multimodal LLMs in dentistry are explored. The authors delineate two primary deployment methodologies: automated dental diagnosis and cross-modal dental diagnosis, elaborating on their prospective utilities. Remarkably, an LLM equipped with a cross-modal encoder can process multi-source data and leverage advanced natural language reasoning for intricate clinical tasks. Beyond the realm of vision-language integration, they underscore the significance of a patient’s voice in medical diagnoses, in conjunction with imaging and dialogues. They illustrate how waveforms and spectrograms from distinct patients could be ingested by pre-trained LLMs like GPT-4 to diagnose potential ailments and gauge their severity. Here, audio data serves a dual purpose: detecting voice anomalies and comprehending patient narratives. In voice anomaly detection, the system captures patient voice inputs, generates waveforms and spectrograms, and subsequently conducts amplitude and frequency analyses. For narrative understanding, patient accounts are transcribed into text via speech recognition technologies. Essential information, such as described symptoms, can then be distilled and organized into concise reports or bullet points for clinician reference.

Furthermore, HeLM~\cite{Belyaeva2023MultimodalLF} demonstrates the value of multimodal LLMs in delivering personalized healthcare. Designed specifically to process high-dimensional clinical data for disease risk assessment, HeLM employs specialized encoders to convert varied data into the LLM's token embedding space, while simpler tabular data is serialized into textual formats. Impressively, HeLM seamlessly merges both demographic and clinical data, including detailed time-series data, to predict disease risks. Moreover, its exceptional performance in zero-shot and few-shot learning for certain conditions reaffirms the immense foundational knowledge that LLMs can contribute to healthcare.

Notably, though the aforementioned multimodal LLMs show promise in handling multimodal data processing and personal user data, serval challenges remain to be addressed. Initially, a substantial volume of multimodal data, which is currently scarce in healthcare, is essential for training these models. There is also a pressing need for research on converting various modalities into aligned embeddings.

\subsection{Core Methodologies}
Shifting from LLMs to multimodal LLMs (MLLMs) often involves a process termed multimodal instruction tuning~\cite{liu2023visual}. Essentially, this method fine-tunes existing LLMs using datasets structured with interwoven text and other data types~\cite{wei2021finetuned}. As the shift from unimodal to multimodal paradigms requires modifications in both dataset structure and model architecture, this section offers an in-depth exploration of the fundamental methodologies for multimodal data acquisition and modality fusion.

\subsubsection{Multimodal Data Acquisition}
Crafting robust multimodal medical datasets for training MLLMs is a meticulous task that has seen various concerted efforts~\cite{Sun2023PathAsstRP,li2023llava}. In this context, we spotlight the dominant techniques for data collection, which mainly hinge on modifying established benchmark datasets and employing the innovative method of self-instruction.

\paragraph{Adapting Established Datasets}
Although existing medical datasets and benchmarks offer a rich source of high-quality data, they often are not formatted for immediate use. Consequently, many studies have undertaken the task of repurposing these datasets into instruction-oriented configurations~\cite{Sun2023PathAsstRP,zhang2023biomedgpt,Thawakar2023XrayGPTCR,wang2023XrayGLM,tu2023towards}. One noteworthy example is PathAsst~\cite{Sun2023PathAsstRP}, which introduces the PathCap dataset, comprised of image-caption pairs primarily curated from a range of reputable sources, including the PubMed database, medical textbooks, and pathology atlases. A pre-trained classifier sorts out pathological data from PubMed. Subsequently, sub-figures and sub-captions are isolated and refined using ChatGPT, leading to a dataset well-suited for multimodal instruction tuning. These image-caption pairs, along with designed instructions, naturally constitute multimodal inputs and responses, where instructions are selected from a pre-established pool. Concurrently, some research initiatives have opted to design a foundational set of instructions~\cite{li2023llava}, expanding them with the aid of GPT-4 for greater diversity and specificity.

\paragraph{Self-Instruction}
While established datasets are invaluable, they often misalign with real-world scenarios, especially in intricate contexts such as multi-turn conversations. To address this discrepancy, several studies~\cite{li2023llava,Sun2023PathAsstRP,wu2023towards,wu2023pmc} have adopted a self-instruction approach~\cite{wang2022self}. Using this technique, LLMs generate textual data based on a foundational set of hand-annotated seed samples. For instance, LLaVA-Med~\cite{li2023llava} leverages GPT-4~\cite{OpenAI2023GPT4TR} to curate instruction-following data with multi-turn conversations around biomedical images. Given an image caption, GPT-4 generates questions and answers, simulating a conversation as if the model could view the image. This interaction is enriched by integrating sentences from the related PubMed articles and by incorporating carefully curated seed examples~\cite{wang2022self} to guide high-quality conversation generation based on the caption and its context. Similarly, PathAsst~\cite{Sun2023PathAsstRP} prompts GPT-4 to produce conversation-based instruction-following data primarily from image captions.

\subsubsection{Modality Fusion}
Given that unimodal LLMs predominantly process text, bridging the gap between natural language and various modalities becomes essential. Two primary strategies have emerged to address this: one involves converting external modalities into natural language using expert models before feeding it into the LLM, and the other continuously integrates external modalities into the LLM's embedding space.

\paragraph{Leveraging Expert Models}
One common approach utilizes expert models, such as image captioning systems, to translate visual data into textual content~\cite{han2023medalpaca,Sun2023PathAsstRP}. Rather than directly processing multimodal inputs, LLMs interpret the transformed textual representations. Visual Med-Alpaca~\cite{han2023medalpaca} exemplifies this strategy, utilizing multiple medical visual expert systems, Med-GIT~\cite{wang2022git} and De-Plot~\cite{liu2022deplot}, in tandem with an LLM. A classifier discerns which specific captioning model should handle the image data, and the generated text is then integrated with the original textual query, enabling the LLM to craft a pertinent response. However, a notable downside is that such transformations could result in information loss. For instance, the granularity of spatial data in visual content might get oversimplified in the conversion to text, and HeLM~\cite{Belyaeva2023MultimodalLF} has demonstrated that direct text serialization of tabular data does not yield a representation that fully captures the available signal for a complex set of features.

\paragraph{Continuous Injection}
Various studies~\cite{tu2023towards,Belyaeva2023MultimodalLF,Moor2023MedFlamingoAM,luo2023biomedgpt,Thawakar2023XrayGPTCR} have aimed to seamlessly merge multimodal data by continuously injecting it into the embedding space of pre-trained LLMs. For instance, Med-PaLM M~\cite{tu2023towards} integrates visual information using the ViT encoder~\cite{Dosovitskiy2020AnII} with PaLM~\cite{Chowdhery2022PaLMSL}, in a manner that sees the continuous integration of visual data. This creates multimodal sentences, where textual data is interspersed with visual embeddings. A representation of such a sentence might look like: \texttt{Q: What happened between <img\_1> and <img\_2>?}, where \texttt{<img\_$i$>} symbolizes an image's embedding. This approach bypasses the discrete token level, allowing for direct mapping of visual observations into the linguistic embedding space. HeLM~\cite{Belyaeva2023MultimodalLF} employs a similar method, converting diverse data types into token embeddings. Nonetheless, a notable challenge with HeLM is the degradation of conversational competence after fine-tuning, a phenomenon observed in other models as well~\cite{Wang2022PreservingIL}.

In more complex scenarios, like with RadFM~\cite{wu2023towards} that deals with 3D images such as MRI or CT scans, the resulting token sequence from visual encoding can be quite lengthy. Here, a perceiver module~\cite{jaegle2021perceiver} is used to compactly represent visual data. Leveraging this architecture, diverse image sizes can be uniformly represented, facilitating easier fusion. Similarly, Med-Flamingo~\cite{Moor2023MedFlamingoAM} uses this perceiver module to efficiently bridge vision encoders and LLMs, translating varying numbers of visual features into a fixed set of outputs, optimizing computational efficiency.

\section{LLM-Powered Autonomous Agents in Medicine}
\renewcommand{\arraystretch}{1.5}
\begin{table*}[t]
\centering
\small
\begin{tabularx}{\textwidth}{
>{\raggedright\arraybackslash}m{2.7cm} 
>{\raggedright\arraybackslash}m{3cm}
>{\raggedright\arraybackslash}m{3.5cm} 
>{\raggedright\arraybackslash}m{3cm}
>{\raggedright\arraybackslash}m{4cm}}
\toprule[1.5pt]
\textbf{Model} & \textbf{Task} & \textbf{Planning} & \textbf{Memory} & \textbf{External Tool Usage} \\ \hline
AD-AutoGPT~\cite{dai2023ad} & Focus on Alzheimer's Disease in medical research & Utilizes GPT-4~\cite{OpenAI2023GPT4TR} for task decomposition & Stores environmental information and past actions & Employs tools for news searching, summarization, and result visualization \\ \hline
CHA~\cite{abbasian2023conversational} & Personalized healthcare conversations, including stress level assessment & Incorporates ReAct~\cite{yao2022react} for task planning & Retains environmental data and historical actions & Utilizes Google Search, translator, healthcare platforms, and stress estimation models \\ \hline
ImpressionGPT~\cite{ma2023impressiongpt} & Summarization of radiology reports & Engages in self-reflection through iterative prompting optimization & Uses similar existing reports as contextual examples & N/A \\ \hline
PharmacyGPT~\cite{liu2023pharmacygpt} & Prescription of medication plans & Applies self-reflection through iterative prompting optimization & Employs patient clustering to provide relevant examples & N/A \\ \hline
ChatCAD+~\cite{zhao2023chatcad+} & Assists in computer-aided diagnosis & Utilizes dynamic prompting for self-reflection & Gathers related information from professional sources & N/A \\ \hline
PathAsst~\cite{Sun2023PathAsstRP} & Assistance in pathological diagnosis & N/A & Retrieves papers from a local knowledge database & Uses expert models for tasks like image segmentation and detection in pathology \\ \hline
Yuan et al.~\cite{Yuan2023AdvancedPA} & Gastrointestinal cancer management & Decomposes task via templated prompt & Retrieves similar cases from storage of knowledge & N/A \\
\bottomrule[1.5pt]
\end{tabularx}

\caption{A summary of LLM-powered autonomous agents in the field of medicine, detailing their specific tasks, planning mechanisms, memory components, and usage of external tools. ``N/A'' denotes components that are not available.}
\label{tab:agent}
\end{table*}


Although LLMs such as ChatGPT, GPT-4~\cite{OpenAI2023GPT4TR}, and Med-PaLM M~\cite{tu2023towards} have made strides in the medical domain, they primarily focus on conversational elements and basic information retrieval. In addition, specialized multimodal LLMs demand a vast amount of multimodal data for training, which is scarce in healthcare. Consequently, these models tend to be task-specific, and their conversational capabilities are limited to their training topics. They are not yet fully equipped to serve as comprehensive healthcare agents due to hurdles in personalization, updating knowledge, and engaging in autonomous sequential thinking, strategic planning, and complex problem-solving, all of which are imperative for physicians in clinical practice~\cite{abbasian2023conversational}. The development of LLM-driven autonomous agents adept at navigating clinical complexities warrants exploration.

In this section, we offer a concise overview of LLM-powered autonomous agents, outlining their key components and features, as illustrated in \autoref{tab:agent}. We proceed to examine existing studies in this domain, classifying them into two primary categories: those focused on developing holistic AI agents for medical applications and those aimed at enhancing individual functionalities of AI agents in healthcare.

\subsection{Autonomous Agents} Autonomous agents are advanced systems capable of independently accomplishing tasks through self-directed planning and instructions \cite{franklin1996agent}. They have emerged as a promising solution for achieving AGI. Traditional approaches have focused on training agents with limited knowledge in isolated environments, failing to achieve human-level proficiency in open-domain settings \cite{mnih2015human}. However, recent breakthroughs in LLMs have exhibited remarkable reasoning capabilities~\cite{wei2022chain}, leading to a growing trend in leveraging LLMs as central controllers to empower autonomous agents. This trend holds the potential to develop general problem solvers, as demonstrated by proof-of-concept demos like AutoGPT \cite{sig2023autogpt}. For a detailed analysis, readers are referred to the comprehensive survey on LLM-based autonomous agents \cite{wang2023survey,weng2023prompt,xi2023rise}. 

In this framework, the LLM acts as the cognitive core of the system, which is further reinforced with various essential capabilities to effectively execute diverse tasks \cite{weng2023prompt}. These capabilities are fulfilled through multiple modules: profile, memory, planning, and action~\cite{wang2023survey}. Specifically, the profile module aims to determine the role profiles of agents, such as radiologists or programmers, which are typically integrated into the prompt to influence and restrict the behaviors of the LLM \cite{hong2023metagpt,argyle2023out}. The memory module stores environmental information and employs recorded memories to facilitate future actions~\cite{hu2023chatdb}. This enables the agent to accumulate experiences, self-evolve, and exhibit more consistent, logical, and effective behavior~\cite{zhong2023memorybank}. The planning module empowers the agent to decompose complex tasks into simpler subtasks and solve them sequentially~\cite{wei2022chain,yao2023tree,zhou2022least}. Additionally, it enables the agent to engage in self-criticism and self-reflection~\cite{shinn2023reflexion,yao2022react}, learning from past actions and refining themselves to enhance future performance. The action module translates the agent's decisions into specific outcomes by directly interacting with the environment. The action capability is enriched by the agent's skill in utilizing diverse external tools or knowledge sources~\cite{mialon2023augmented,schick2023toolformer}, such as APIs, knowledge bases, and specialized models. These modules are interconnected to establish an LLM-based autonomous agent, where the profiling module influences the memory and planning modules, which, together with the profile, collectively impact the action module~\cite{xi2023rise}.

\subsection{Developing Comprehensive AI Agents in Medicine}
Autonomous agents powered by LLM technology, equipped with advanced language comprehension and reasoning abilities, have had a revolutionary impact on various disciplines. They have been successfully employed as assistants in natural science experiments~\cite{boiko2023emergent,bran2023chemcrow} and software engineering projects~\cite{qian2023communicative,hong2023metagpt}. However, the potential of autonomous agents in the medical field remains largely unexplored~\cite{dai2023ad,abbasian2023conversational}, due to the complex nature of clinical practice. 

AD-AutoGPT~\cite{dai2023ad} has made the first attempt to develop comprehensive AI agents in the medical field, where a specialized AI agent is constructed to autonomously collect, process, and analyze complex health narratives related to Alzheimer's Disease based on textual prompts provided by the user. This agent leverages ChatGPT~\cite{ouyang2022training} or GPT-4~\cite{OpenAI2023GPT4TR} for task decomposition and augments it with a library of instructions that includes customized tools such as news search, summarization, and result visualization. Moreover, it incorporates specific prompting mechanisms to enhance the efficiency of retrieving AD-related information and employs a tailored spatiotemporal information extraction functionality. This pipeline revolutionizes the conventional labor-intensive data analysis approach into a prompt-based automated framework, establishing a solid foundation for subsequent AI-assisted public health research.

Abbasian et al.~\cite{abbasian2023conversational} put forth an innovative system, the Conversational Health Agent (CHA), leveraging LLMs to revolutionize personal healthcare services through empathetic dialogue and sophisticated processing of multimodal data. This comprehensive system navigates through a series of pivotal steps, initiating the extraction of user queries from conventional multimodal conversations and transforming them into a structured sequence of executable actions to craft the final response. It then demonstrates its problem-solving prowess by tapping into LLMs as a robust knowledge base for a variety of healthcare tasks. Concurrently, it meticulously retrieves the latest and most relevant healthcare information from reputable published sources, aligning it with the user's specific inquiries. The CHA extends its capabilities by forming connections with diverse external health platforms to acquire up-to-date personal user data. When necessary, it delves into multimodal data analysis, utilizing state-of-the-art external machine-learning healthcare tools. Culminating its processes, the system synthesizes all accumulated information to generate responses that are both tailored to the individual user and reflect the most current knowledge, ensuring transparent communication and providing elucidations on the reasoning and reliability of its approach upon user request. The framework proves its mettle by adeptly handling intricate, multi-step health tasks, such as assessing user stress levels, requiring a nuanced blend of personalization, multimodal data analysis, and extensive health knowledge retrieval.

\subsection{Fulfilling Individual Functions of AI Agents}
While comprehensive AI agents tailored for the medical field remain relatively rare, previous research~\cite{ma2023impressiongpt,liu2023pharmacygpt,zhao2023chatcad+,Sun2023PathAsstRP,Yuan2023AdvancedPA} that addresses and fulfills individual modules or functions can be considered as initial iterations of AI agents in medicine, illuminating the potential of such agents. 

ImpressionGPT~\cite{ma2023impressiongpt} introduces the utilization of LLMs for the purpose of summarizing radiology reports. It presents a dynamic prompt approach that employs similarity search techniques to incorporate existing reports that are semantically and clinically similar. These similar reports are then used as demonstrations to assist ChatGPT~\cite{ouyang2022training} in learning the text descriptions and summarizations of comparable imaging manifestations within a dynamic context. This enables the model to acquire contextual knowledge from analogous instances in the available data, leveraging the in-context learning capabilities of LLMs~\cite{brown2020language} in a manner that aligns with the memory module of autonomous agents. Furthermore, an iterative optimization algorithm is devised to automatically evaluate the generated results and formulate corresponding instruction prompts. This iterative process enhances the model by facilitating self-reflection, similar to the self-reflection capability observed in autonomous agents~\cite{shinn2023reflexion,yao2022react}. 
In addition, PharmacyGPT~\cite{liu2023pharmacygpt} extends this framework to address various clinically significant challenges in the pharmacy domain, including patient outcome studies, the generation of AI-based medication prescriptions, and the interpretable clustering analysis of patients, showcasing the versatile potential of autonomous agents in medicine. 

Additionally, ChatCAD+~\cite{zhao2023chatcad+}, a comprehensive and dependable computer-assisted diagnosis system, is capable of analyzing medical images across a wide range of domains and utilizing up-to-date medical information from reputable sources to offer reliable advice. Specifically, given the input medical image, the system incorporates CLIP~\cite{radford2021learning} as a domain identifier to select an appropriate model to generate textual descriptions that sufficiently characterize image features. Rather than providing diagnostic advice directly, ChatCAD+ first retrieves pertinent knowledge from professional sources such as Mayo Clinic. This curated information bolsters the LLM's knowledge pool, amplifying the reliability of its diagnostic advice. Additionally, ChatCAD+ integrates a template retrieval mechanism, used in ImpressionGPT~\cite{ma2023impressiongpt}, to further improve the report generation performance. Besides, PathAsst system~\cite{Sun2023PathAsstRP} is fine-tuned to have the capabilities of invoking external pathological models and retrieving relevant information from an extensive paper database, making the system capable of handling more complicated tasks and elevating the precision and thoroughness of the responses.

\section{Evaluation}
Given the rapid advancements in the capabilities of LLMs, their integration into critical domains like medicine necessitates a stringent evaluation. These models, while technological marvels, have profound implications on medical decision-making, patient care, and the broader healthcare landscape. As we delve deeper into their technical prowess and applications, the need to ascertain their performance and reliability in medical settings is paramount.

Ensuring the efficacy and safety of LLMs in medicine is vital. An established evaluation framework serves two main purposes: it safeguards against inaccuracies or misjudgments that might have negative consequences in the high-stakes realm of healthcare and provides clear benchmarks and metrics that drive ongoing research and development. Within this framework, evaluating LLMs can be stratified into closed-set and open-set categories based on question types. This distinction is important as it illuminates the model's capabilities in predefined tasks and their adaptability to real-world, unpredictable medical inquiries.
\subsection{Closed-Set Evaluation}
Closed-set questions come with predefined and limited answer options. Their evaluation often uses benchmark-adapted datasets, with performance metrics derived from these standards. For example, LLaVA-Med~\cite{li2023llava} measures accuracy for closed-set questions using datasets such as VQA-RAD~\cite{lau2018dataset} and SLAKE~\cite{liu2021slake}. Evaluation settings typically utilize either a zero-shot approach or finetuning. The former takes a range of datasets encompassing various tasks, dividing them into 'held-in' (used for training) and 'held-out' sets (used for testing). After training on the 'held-in' sets, performance on unseen datasets or tasks is measured. In contrast, finetuning is more common in domain-specific task evaluations, as demonstrated by LLaVA-Med's results on biomedical VQA~\cite{lau2018dataset,liu2021slake}.

Despite their usefulness, these evaluations often cover only a limited set of tasks or datasets, lacking a broad quantitative comparison. Recent efforts have sought to bridge this gap, like Med-PaLM's introduction of MultiMedQA~\cite{Singhal2022LargeLM}, which consolidates six medical Q\&A datasets, and the addition of another dataset from online medical queries. Another significant contribution is MultiMedBench by Med-PaLM M~\cite{tu2023towards}, a comprehensive benchmark for biomedical tasks. RadFM's RadBench~\cite{wu2023towards} caters specifically to radiology.

\subsection{Open-set Evaluation}
Open-set questions allow for a wider range of responses, making LLMs function similarly to chatbots in this context. Given the diverse content, evaluating these responses is multifaceted. Metrics cover standard measures, expert reviews, model scores, and other unique aspects. The model should prioritize clinical relevance, ensuring its information directly influences patient care. Accuracy, safety, interpretability, ethical considerations, and scalability are also of paramount importance, ensuring the model's predictions are trustworthy and widely applicable.

\paragraph{Standard Metrics} Standardized metrics established in the NLP community are often employed to evaluate LLM linguistic outputs. These include F1 score~\cite{wu2023towards}, accuracy~\cite{wu2023towards}, precision~\cite{wu2023towards}, recall~\cite{li2023llava}, BLEU~\cite{Papineni2002BleuAM}, METEOR~\cite{banerjee2005meteor}, and ROUGE score~\cite{Lin2004ROUGEAP}. For instance, BLEU evaluates word and phrase overlaps between a model's output and a reference, while METEOR measures lexical and semantic similarities between the generated summary and the reference. These metrics, which range from 0.0 to 1.0, reflect how closely generated outputs match reference answers.

\paragraph{Expert Evaluation} In the healthcare domain, model evaluation goes beyond standard metrics like BLEU and ROUGE, given the evident discrepancies when human evaluations depart from automated benchmarks~\cite{Yuan2023AdvancedPA,singhal2023large,tu2023towards}. Med-PaLM's findings underscored that even top-performing models, such as Flan-PaLM~\cite{Singhal2022LargeLM}, might not always align with clinicians' preferences. The introduction of clinical radiology-tailored metrics, accompanied by expert assessments on aspects like clinical relevance, offers a more grounded evaluation. Both Yuan et al.\cite{Yuan2023AdvancedPA} and Xu et al.\cite{xu2023medgpteval} have developed metrics based on clinical evaluations to further refine model assessment. The robust evaluation process begins with pilot studies, followed by expert peer reviews, culminating in real-world clinical tests. This comprehensive framework ensures not only the model's accuracy but also its applicability and safety. Once thoroughly vetted, such models can gradually integrate into clinical workflows, aiding professionals in diagnostics, treatment suggestions, and more.

\paragraph{Model Scoring} To address the resource-intensive nature of manual assessments, researchers~\cite{liu2023gpteval,shi2023llm,fu2023gptscore,chen2023exploring,chiang2023can,zhang2023huatuogpt} are exploring LLM-based scoring systems like GPT-Eval~\cite{liu2023gpteval} and LLM-Mini-CEX~\cite{shi2023llm}. These systems employ model-centric strategies, wherein one LLM, usually GPT-4~\cite{OpenAI2023GPT4TR}, evaluates another one's medical dialogues. For instance, GPT-Eval~\cite{liu2023gpteval} provides a methodology where the task and criteria are fed into an LLM, leading to a series of evaluation steps that another LLM uses for assessment. LLM-Mini-CEX~\cite{shi2023llm} offers a unique LLM-tailored criterion, streamlining evaluation of diagnostic abilities by automating interactions using a patient simulator and ChatGPT. However, these methods face challenges related to transparency, accuracy, and sometimes limited diagnostic performance, as noted by Shi et al.~\cite{shi2023llm}.

\paragraph{Other Aspects} There are also evaluations focusing on unique LLM characteristics~\cite{chang2023survey}, such as faithfulness~\cite{xie2023faithful}, hallucination~\cite{umapathi2023med}, safety~\cite{zhang2023safetybench}, and robustness against adversarial interventions~\cite{wang2021adversarial}. For example, to address the challenges posed by hallucinations in LLMs, particularly in the context of the medical domain, Med-HALT~\cite{umapathi2023med} provides a diverse multinational dataset derived from medical examinations across various countries and includes multiple innovative testing modalities, especially reasoning hallucination tests.
\\\\
\indent In summary, as LLMs show promising advancements in the medical domain, rigorous and comprehensive evaluations are crucial. A combination of automated metrics, expert evaluations, and real-world testing ensures the models' efficacy and safety. As technology and medicine further intertwine, evaluation frameworks must evolve accordingly to ensure the best patient outcomes. Once a model passes this framework, its gradual integration into clinical workflows can commence, starting with tasks like summarizing medical records or aiding in diagnostics, but always under medical professionals' supervision.
\section{Discussion}
In this review, we have meticulously navigated through the multifaceted landscape of Large Language Models (LLMs) in the medical domain, illuminating their promising potential. We initiated our exploration by delving into the foundational applications of LLMs in medicine, emphasizing text-based interactions and distinguishing between general-purpose and specialized medical LLMs. Recognizing the inherent multimodality of the medical field, our discussion transitioned to multimodal LLMs, highlighting their capability to integrate diverse data types and thereby augment diagnostic accuracy. Despite these advancements, we acknowledged the persisting challenges such as the need for personalized responses, maintaining currency with the latest medical knowledge, and navigating complex problem-solving scenarios—skills that are indispensable for clinicians in clinical settings. In response to these challenges, we scrutinized the emerging role of LLM-powered autonomous agents in medicine, categorizing their applications and summarizing prevailing evaluation methodologies. 

Through this extensive analysis, we aimed to provide a balanced and nuanced perspective on the current state of LLMs in medicine. In the contemporary landscape of LLM development, there is a discernible trend toward harnessing LLMs specifically for the medical domain. While general-purpose LLMs exhibit remarkable proficiency, our observations suggest a strategic advantage in not directly fine-tuning them on specialized, long-tailed medical data. Instead, employing highly specialized expert models to handle such nuanced data, followed by storing the processed information in vector databases~\cite{zhao2023chatcad+,Sun2023PathAsstRP}, emerges as a promising paradigm. In practice, querying this database can offer accurate and domain-specific insights. This approach not only presents a potential solution to the "hallucination" phenomenon, where the models may fabricate inconsistent or outright false information, but also paves the way for integration within an autonomous agent-based system, hinting at a comprehensive medical support system for the future. Amidst these advancements, we emphasize the potential of LLMs to revolutionize medical practice while underscoring the imperative for ethical vigilance and continuous scrutiny.

The integration of LLMs in the medical field necessitates a nuanced evaluation from both technological and medical standpoints. From a technical perspective, the proficiency of LLMs in parsing and generating complex, nuanced language is crucial, particularly in understanding and formulating medical terminology, patient narratives, and intricate case details. However, the efficacy of these models hinges on their ability to handle sensitive information ethically, maintain patient confidentiality, and navigate the consequences of misinformation, requiring strict accuracy benchmarks. For medical professionals, the assessment revolves around the practical applicability of LLMs: do they enhance diagnostic accuracy, improve patient communication, and aid in advanced research and treatment methodologies without compromising professional responsibilities or patient trust? Ultimately, the symbiotic evaluation seeks to ensure that LLMs not only exhibit technical excellence but also adhere to the rigorous ethical and professional standards indispensable in healthcare. 

Although LLMs possess remarkable intellectual abilities, they inherently face certain constraints and are prone to producing inaccurate or possibly damaging content. Their understanding of context is limited, which is particularly problematic in medical settings where precision is crucial. The models' potential to misinterpret or oversimplify complex medical information necessitates rigorous human oversight. Moreover, LLMs' knowledge is constrained to their training data~\cite{arora2023promise,Sallam2023.02.19.23286155,biswas2023chatgpt,kothari2023chatgpt}, potentially leading to outdated or biased responses, and their inability to ask follow-up questions can be a significant drawback in patient care. 

Ethical dilemmas significantly pervade the deployment of LLMs in the medical sphere. These models, by inheriting biases from their vast swathes of training data, are at risk of producing responses that are prejudiced or discriminatory, thereby challenging the principles of fairness and equity in healthcare. The 
"hallucination" phenomenon pose a substantial risk, particularly in medical settings where accuracy is paramount~\cite{shen2023chatgpt}. Moreover, the legal landscape~\cite{biswas2023chatgpt} surrounding the use of LLMs in medicine is intricate and fraught with potential pitfalls, necessitating meticulous attention. Issues such as copyright infringement, plagiarism~\cite{Sallam2023.02.19.23286155}, defamation, and breaches of privacy are prominent concerns that must be proactively addressed to safeguard against legal repercussions and uphold ethical standards. Complicating these issues is the models' inherent complexity and the opaqueness of their internal mechanisms. This lack of transparency hinders the ability to decipher how specific outputs are generated, leading to potential trust and accountability issues~\cite{Sallam2023.02.19.23286155, ufuk2023role}.

Despite these challenges, LLMs hold the potential to transform healthcare by enhancing research, improving operational efficiency, and aiding in decision-making. For successful integration and adoption in healthcare, it is imperative to address these limitations and ensure ethical, reliable, and safe applications. Future directions should focus on developing frameworks that recognize and mitigate these constraints, promoting a responsible and informed use of LLMs in medicine.

\section{Conclusion}
In conclusion, this review offers a comprehensive analysis of the transformative potential of large language models (LLMs) in modern medicine. It demonstrates the fundamental applications of general-purpose and specialized LLMs in areas like knowledge retrieval, research support, workflow automation, and diagnostic assistance. Recognizing the multimodal nature of medicine, the review explores multimodal LLMs and their ability to process diverse data types including medical imaging and EHRs to enhance diagnostic accuracy. To address limitations of LLMs regarding personalization and complex clinical reasoning, the emerging role of LLM-powered autonomous agents in medicine is discussed. The review also summarizes methodologies for evaluating the reliability and safety of LLMs in medical contexts.

Overall, LLMs hold remarkable promise in medicine but require continuous optimization and ethical oversight before effective integration into clinical practice. Key challenges highlighted include data limitations, reasoning gaps, potential biases, and transparency issues. Future priorities should focus on developing frameworks to identify and mitigate LLM limitations, guiding responsible and informed applications in healthcare. With prudent progress, LLMs can transform modern medicine through enhanced knowledge consolidation, personalized care, accelerated research, and augmented clinical decision-making. But ultimately, human expertise, ethics and oversight will remain indispensable in delivering compassionate, high-quality and equitable healthcare.

\section*{Acknowledgements}
This work was supported by the National Natural Science Foundation of China (91959205 to L.S., U22A20327 to L.S., 82203881 to Y.C., 82272627 to XT.Z., 7232018 to Y.S., 12090022 to B.D., 11831002 to B.D., 81801778 to L.Z.), Beijing Natural Science Foundation (7222021 to Y.C., Z200015 to XT.Z.), Beijing Hospitals Authority Youth Programme (QML20231115 to Y.C.), Clinical Medicine Plus X-Young Scholars Project of Peking University (PKU2023LCXQ041 to Y.C. and L.Z.), Guangdong Provincial Key Laboratory of Precision Medicine for Gastrointestinal Cancer (2020B121201004).

\bibliographystyle{ieeetr}
\bibliography{sample}
\end{document}